\documentclass[lnbip]{svmultln}
\usepackage{makeidx}  
\usepackage{url}

\usepackage{mathptmx}
\usepackage[most]{tcolorbox}
\usepackage{makeidx}
\usepackage[english]{babel}
\usepackage{amsmath}
\usepackage{amsfonts}
\usepackage{amssymb}
\usepackage{graphicx}
\usepackage{xspace}
\usepackage{url}
\usepackage{algorithm}
\usepackage{algpseudocode}
\usepackage{booktabs}
\usepackage{multirow}
\usepackage[labelsep=quad,indention=10pt]{subfig}
\usepackage{bbding}
\usepackage{todonotes}
\usepackage{adjustbox}
\usepackage{rotating}
\usepackage{paralist}
\usepackage{float}

\usepackage[normalem]{ulem}
\newif\ifrevisionmode
\revisionmodefalse

\ifrevisionmode
    
    \newcommand{\remove}[1] { {\color{red}\sout{#1}} }
    
\else
    
    \newcommand{\remove}[1]{}
    
\fi

\usepackage[normalem]{ulem}
\usepackage{graphicx}
\newcommand{\sysfont}{\textit}
\newcommand{\dlv}{\sysfont{DLV}\xspace}
\newcommand{\system}{\sysfont{I-DLV-sr}\xspace}
\newcommand{\idlv}{{{\small $I$}-}\dlv}
\newcommand{\iidlv}{{{\small $\cal I$}$^2$-}\dlv}
\newcommand{\quot}[1]{``#1"}

\newcommand{\derives}{\mbox{\,:\hspace{0.1em}\texttt{-}}\,\xspace}

\begin{document}
\mainmatter              
%
\title{A Formal Comparison between Datalog-based Languages for Stream Reasoning \\ (extended version)}
\titlerunning{A Formal Comparison between Datalog-based Languages for Stream Reasoning}  
%
\author{Nicola Leone\inst{1} \and Marco Manna\inst{1} \and Maria Concetta Morelli \inst{2}\and Simona Perri \inst{1}}
\authorrunning{Nicola Leone et al.} 
%
\tocauthor{Nicola Leone, Marco Manna, Maria Concetta Morelli, Simona Perri}
\institute{Department of Mathematics and Computer Science, University of Calabria, Rende, Italy\\
\email{name.surname@unical.it},
\and
Department of Mathematics and Computer Science, University of Calabria, Rende, Italy\\
\email{maria.morelli@unical.it}}

\maketitle              

\begin{abstract}
The paper investigates the relative expressiveness of two logic-based languages for reasoning over streams, namely 
\textit{LARS} Programs --- the language of the Logic-based framework for Analytic Reasoning over Streams called \textit{LARS} --- and $\textit{LDSR}$ --- the language of the recent extension of the \idlv system for stream reasoning called \system. Although these two languages build over Datalog, they do differ both in syntax and semantics. To reconcile their expressive capabilities for stream reasoning, we define a comparison framework that allows us to show that, without any restrictions, the two languages are incomparable and to identify fragments of each language that can be expressed via the other one.
\end{abstract}

\section{Introduction}

Stream Reasoning (SR)~\cite{DBLP:journals/datasci/DellAglioVHB17} is a recently emerged research area that consists in the application of inference techniques to heterogeneous and highly dynamic streaming data.
Stream reasoning capabilities are nowadays a key requirement for deploying effective applications in several real-world domains, such as IoT, Smart Cities, Emergency Management.
Different SR approaches have been proposed in contexts such as Complex Event Processing, Semantic Web and Knowledge Representation and Reasoning (KRR)~\cite{DBLP:journals/ijsc/BarbieriBCVG10,hoeksema2011high,DBLP:conf/lpnmr/PhamAM19,DBLP:conf/semweb/PhuocDPH11}. 
In the KRR research field, the Answer Set Programming (ASP) declarative formalism~\cite{DBLP:journals/cacm/BrewkaET11,DBLP:conf/ijcai/GebserLMPRS18} has been acknowledged as a particularly attractive basis for SR~\cite{DBLP:journals/datasci/DellAglioVHB17} and a number of SR solutions relying on ASP have been recently proposed ~\cite{beck2018lars,bazoobandi2017expressive,beck2018ticker,eiter2019distributed,DBLP:conf/bigdataconf/RenCNX18,DBLP:conf/rr/MileoAPH13,DBLP:conf/lpnmr/PhamAM19,DBLP:conf/ai/DoLL11,DBLP:conf/lpnmr/GebserGKS11,DBLP:journals/tplp/CalimeriMMMPZ21}. 
Among these,  \system~\cite{DBLP:journals/tplp/CalimeriMMMPZ21} is a SR system that efficiently scale over real-world application domains thanks to a proper integration of the well-established stream processor \textit{Apache Flink}~\cite{carbone2015apache} and the
incremental ASP reasoner \iidlv~\cite{DBLP:journals/tplp/IanniPZ20}.
Its input language, called \textit{LDSR} (the Language of \idlv for Stream Reasoning) inherits the highly declarative nature and ease of use from ASP, while being extended with new constructs that are relevant for practical SR scenarios.

Although $\textit{LDSR}$ enjoys valuable knowledge modelling capabilities together with an efficient and effective reasoner, it would be desirable to formally investigate its expressive power. This is exactly the mission of the present paper.
To this aim, the prime objective is to compare $\textit{LDSR}$ with one of the most famous and well-studied formalisms for reasoning over streams that goes under the name of \textit{LARS Programs} (the language of \textit{LARS}), where \textit{LARS} is the Logic-based framework for Analytic Reasoning over Streams~\cite{beck2018lars}.
More precisely, the paper investigates the relative expressiveness of $\textit{LDSR}$ and \textit{LARS} Programs. 
Despite the fact that these languages build over Datalog
and represent the information via sets of ASP ground predicate atoms associated with different time points, unfortunately they overall differ both in syntax and semantics. 
In particular, $\textit{LDSR}$ associates information that is true at every time point with standard ASP facts, whereas \textit{LARS} Programs involve background atoms whose truth is not directly associated with all the time points.
Moreover, given an input stream, $\textit{LDSR}$ returns a single set of information related to the most recent time point (streaming model), whereas \textit{LARS} Programs associate, for each different time point of the stream, another stream called answer stream at that time point.

To reconcile the expressive capabilities for stream reasoning of $\textit{LDSR}$ and \textit{LARS} Programs, we first define a comparison framework that allows to understand in which cases, starting from the same input stream, both languages may produce the same output. 
Inside this framework, we identify three output profiles ---called {\em atomic}, {\em bound}, and {\em full}--- that fix the form of the output stream. 
Without any restriction on the two languages, the paper shows that they are incomparable under all the output profiles. 
Eventually, for each output profile, the paper isolates large fragments of each of the two languages that can be expressed via the other one.

\section{Preliminaries}\label{sec:preliminaries}
We assume to have finite sets $\cal{V}$, $\cal{C}$, $\cal{P}$ and $\cal{U}$ consisting of \emph{variables}, \emph{constants}, \emph{predicate names} and $time$ $
variables$ respectively; we constrain $\cal{V}$ and $\cal{C}$ to be disjoint.
A \emph{term} is either a variable in $\cal{V}$ or a constant in $\cal{C}$.
A \emph{predicate atom} has the form $p(t_1,\dots,t_n)$, where $p\in {\cal P}$ is a predicate name, $t_1,\dots,t_n$ are terms and $n\geq0$ is the arity of the predicate atom; a predicate atom $p()$ of arity $0$ can be also denoted by $p$. 
A predicate atom is \emph{ground} if none of its terms is a variable. 
We denote by $G$ the set of all ground predicate atoms constructible from predicate names in ${\cal P}$ and constants in $\cal{C}$. We divide the set of predicates ${\cal P}$ into two disjoint subsets, namely the extensional predicates ${\cal P}^{\varepsilon}$ and the intensional predicates ${\cal P}^{I}$. Extensional predicates are further partitioned into ${\cal P}_{B}^{\varepsilon}$ for background data and  ${\cal P}_{S}^{\varepsilon}$ for data streams. The mentioned partitions are analogously defined for ground atoms $G^{I}$,$G_{B}^{\varepsilon}$ and $G_{S}^{\varepsilon}$. In what follows we will introduce different types of constructs peculiar to the two languages considered in this paper. In both, given a set of constructs $C$, we will denote by $pred(C)$ the set of predicates appearing in $C$.

A \emph{stream $\Sigma$} is a sequence of sets of ground predicate atoms $\langle S_0,\dots, S_n\rangle$ such that for $\;0\leq i \leq n$, $S_i\subseteq G$. 
Each natural number $i$ is called \emph{time point}. 
A ground predicate atom $a\in S_i$ is true at the $i$-th time point. Given a value $m\in \mathbb{N}$ s.t. $0\leq m\leq n $, we define the \emph{restriction} of $\Sigma$ to $m$ the stream $\langle S_0,\dots,S_m\rangle$ denoted with $\Sigma_{|_m}$. Moreover, let $F\subseteq \cal{P}$ 
we indicate with $\Sigma_{|_F}$ the stream $\langle S'_0,\dots, S'_n\rangle$ with $S'_i=\bigcup_{\{a \in S_i|pred(a)\in F\}}a$ for each $i\in \{0,\dots,n\}$. 
%
%
A subset of a stream $\Sigma=\langle S_0,\dots, S_n\rangle$ is a stream $\Sigma'=\langle S'_0,\dots, S'_n\rangle$ such that $S'_i=\emptyset$ for each $i\not\in t(\Sigma')$ where $t(\Sigma')$ is a subset of consecutive numbers of the set $\{0,\dots,n\}$ and for each $i\in t(\Sigma')$ $S'_i\subseteq S_i$.
For a stream $\Sigma=\langle S_0,\dots, S_n\rangle$, a {\em backward observation} identifies ground predicate atoms that are true at some time points preceding the $n$-th time point.
%
More formally, given a stream $\Sigma=\langle S_0,\dots, S_n\rangle$ and a set of numbers $D$ $\subset \mathbb{N}$, we define the \emph{backward observation of $\Sigma$ w.r.t. $D$} as the family of sets
$\{S_i\ \vert\ i=n-d \text{ with } d\in D \wedge i\geq0\}$, and we denote it as $O(\Sigma, D)$.
Given $w\in\mathbb{N}$, a backward observation of $\Sigma$ w.r.t. $\{0,\dots,w\}$ is called \emph{window}.

\subsection{\textit{LARS} syntax and semantics}\label{sec:LARS preliminaries}
A {\em window function} is a function $f_w$ that returns, given a stream $\Sigma$ and a time point $t\in \{1,...,n\}$, a substream of $\Sigma$. We consider only time-based window functions, which select all the atoms appearing in the last $w$ time points, to which a window is trivially associated according to the definition above. 
Given a predicate atom $a$, a term $t\in N \cup \cal{U}$ and a window function $f_w$, formulas $\alpha$ are defined by the following grammar:

\[
\alpha ::= a \ | \ \neg\alpha \ | \ \alpha \wedge\alpha \ | \ \alpha \vee \alpha \ | \ \alpha \rightarrow \alpha \ | \ \diamond\alpha  \ | \ \square\alpha \ | \ @_t \alpha \ | \ \boxplus^{f_w} \alpha \ | \ \triangleright \alpha.    
\]

\noindent A \textit{LARS} program P is a set of rules of the form
$\alpha \leftarrow \beta_1,\dots, \beta_n$, where $\alpha ,\beta_1,\dots, \beta_n$ are formulas. Given a rule $r$, we call $\alpha$ the \emph{head} of $r$, denoted with $H(r)$, and we call the conjunction $\beta_1\wedge\dots\wedge \beta_n$ the \emph{body} of $r$, denoted  with $B(r)$. A ground formula can be satisfied at a time point $t$ in a $structure$ that is a triple $M=(\Sigma,W,B)$ where $\Sigma$ is a stream, $W$ is a set of window functions and $B\subseteq G_{B}^{\varepsilon}$. Let $\Sigma' \subseteq \Sigma$, we start defining the entailment relation $\Vdash$ between $(M,\Sigma',t)$ and formulas:

\begin{itemize}
\item $M,\Sigma', t \Vdash  a$ iff $a \in S_t$ or $a \in B$
\item $M,\Sigma', t \Vdash \neg \alpha$ iff $M, t \not\Vdash \alpha$
\item $M,\Sigma', t \Vdash \alpha\wedge \beta$ iff $M, t \Vdash \alpha$ and $M, t \Vdash \beta$
\item $M,\Sigma', t \Vdash \alpha\vee \beta$ iff $M, t  \Vdash \alpha$ or $M, t \Vdash \beta$
\item $M, \Sigma', t \Vdash \alpha\rightarrow \beta$ iff $M, t \not\Vdash \alpha$ or $M, t \Vdash \beta$
\item $M, \Sigma', t \Vdash \diamond \alpha$ iff $M, \Sigma', t \Vdash \alpha$ for some $t' \in t(\Sigma')$
\item $M, \Sigma', t  \Vdash \square \alpha$ iff $M, \Sigma', t \Vdash \alpha$ for all $t' \in t(\Sigma')$ ,
\item $M, \Sigma', t \Vdash @_{t'} \alpha$ iff $M, \Sigma', t' \Vdash \alpha$ and $t'$ $ \in t(\Sigma')$ ,
\item $M, \Sigma', t \Vdash \boxplus^{f_w} \alpha$, iff $M, \Sigma'', t \Vdash \alpha$ where $\Sigma'' = f_w(\Sigma', t),$
\item  $M, \Sigma', t \Vdash \triangleright \alpha$ iff $M, \Sigma, t \Vdash \alpha$.
\end{itemize}

\noindent The structure $M=(\Sigma,W,B)$ satisfies $\alpha$ at time $t$ ($M,t\models \alpha)$ if $M, \Sigma', t \Vdash \alpha$. 
Given a ground \textit{LARS} program $P$, a stream $\Sigma$ and a structure $M$ we say that: $(i)$ $M$ is a  {\em model} of rule $r\in P$ for $I$ at time $t$, denoted $M, t\models r$, if $M,t\models B(r) \rightarrow H(r)$; $(ii)$ $M$ is a {\em model} of $P$ for $I$ at time $t$, denoted $M,t\models P$, if $M, t\models r$ for all rules $r\in P$; $(iii)$
 $M$ is a {\em minimal model}, if no model $M'=(\Sigma',W,B)$ of $P$ for $I$ at time $t$ exists such that $\Sigma'\subset \Sigma $ and $t(\Sigma)=t(\Sigma')$; and
$(iv)$ The {\em reduct} of a program $P$ with respect to $M$ at time $t$ is defined by  $P^{M,t}=\{r\in P|M,t\models B(r)\}$.
Fixed an input stream $I$, contains only atoms belong to $G_{S}^{\varepsilon}$, we call $interpretation \; stream$ $for \;I$ any stream $\Sigma$ such that all atoms that occur in $\Sigma$ but not in $I$ have intensional predicates. An interpretation stream $\Sigma$ for a stream $I$ is an $answer \;stream$ of a program $P \; for \; I \; at \; t$, if $M=(\Sigma,W,B)$ is a $\subseteq $-minimal model of the reduct $P^{M,t}$ for $I$ at time $t$.
The semantics of the non-ground programs is given by the answer streams of according groundings, obtained by replacing variables with constants from $\cal{C}$, respectively time points from $t(\Sigma)$, in all possible ways.
We consider \textit{LARS} programs with a single answer stream for each time point, denoted with $\textit{LARS}_{\textit{D}}$, and we indicate the single answer stream of $P$ for $I$ at $t$ with $AS(P,I,t)$.

\subsection{$\textit{LDSR}$ syntax and semantics} \label{sec:LDSR preliminaries}
Given a predicate atom $a$, a term $c\in  \cal{C}\cap \mathbb{N^+}$, a \emph{counting term} $t\in (\cal{C}\cap \mathbb{N^+})\cup V$, and a finite non-empty set $D = \{d_1,\dots,d_m\} \subset\mathbb{N}$, we define three types of {\em streaming atoms}: 
\[
a  \;\mathbf{at \;least}\; c \;\mathbf{in} \; \{ d_1,\dots,d_m\}
\  \ | \ \ 
a  \;\mathbf{always \; in} \; \{d_1,\dots,d_m\}
\  \ | \ \ 
a  \;\mathbf{count}\; t \;\mathbf{in} \; \{ d_1,\dots,d_m\}  
\]
\noindent In particular, if $D$ is of the form $\{0,...,w\}$, then this set can be alternatively written as $[w]$ inside streaming atoms; also
we may write $a$ in place of  $a  \;\mathbf{at \;least}\; 1 \;\mathbf{in} \; [0]$.
A streaming atom $\alpha$ (resp., $\mathtt{not}\ \alpha$) is said to be a \emph{positive streaming literal} (resp., \emph{negative streaming literal}), where $\mathtt{not}$ denotes \emph{negation as failure}. %
A streaming literal is said to be \textit{harmless} if it has form 
$a \;\mathbf{at \;least}\; c \;\mathbf{in} \;  D$ or 
$a \;\mathbf{always \; in}$ $\;  D$; otherwise, it is said to be \textit{non-harmless}.
A streaming literal is said to be {\em ground} if none of its terms is a variable. 

A {\em rule} is a formula of form
$(1)\ \ a \derives\; l_1,\dots,l_b.$\ \  or $(2)\ \ \mathbf{\#temp} \; a \derives\; l_1,\dots,l_b.$, where $a$ is a predicate atom, $b\ge0$ and $l_1,\dots,l_b$ represent a conjunction of literals (streaming literals or other literals defined in the ASP-Core-2 standard~\cite{DBLP:journals/tplp/CalimeriFGIKKLM20}).

For a rule $r$, we say that the \emph{head} of $r$ is the set $H(r) = \{a\}$, whereas the set $B(r) = \{l_1,\dots,l_b\}$ is referred to as the \emph{body} of $r$. 
A rule $r$ is \textit{safe} if all variables in $H(r)$ or in a negative streaming literal of $B(r)$ also appear in a positive streaming literal of $B(r)$.

A program $P$ is a finite set of safe rules. We denote with $\mathit{form}_{(1)}(P)$ the set of rules of $P$ of form (1) and with $\mathit{form}_{(2)}(P)$ the set of rules of $P$ of form (2). 

A program $P$ is \emph{stratified} if there is a partition of disjoint sets of rules $P=\Pi_1\cup \dots\cup\Pi_k$ (called strata) such that for $i\in\{1,\dots,k\}$ both these conditions hold:
     ($i$) for each harmless literal in the body of a rule in $\Pi_i$ with predicate $p$, $\{r \in P \vert H(r)=$ $\{p(t_1,\ldots,t_n)\}\}\subseteq \bigcup_{j=1}^{i} \Pi_j$;
    ($ii$) for each non-harmless literal in the body of a rule in $\Pi_i$ with predicate $p$, $\{r \in P \vert H(r)=$ $\{p(t_1,\ldots,t_n)\}\}\subseteq \bigcup_{j=1}^{i-1} \Pi_j$.
We call $\Pi_1,\dots,\Pi_k$ a \emph{stratification} for $P$ and $P$ is stratified by $\Pi_1,\dots,\Pi_k$.
An \textit{LDSR} program is a program being also stratified.

A backward observation allows to define the truth of a ground streaming literal at a given time point.
Given a stream $\Sigma=\langle S_0,\dots, S_n\rangle$, $D=\{d_1,\dots,d_m\}\subset \mathbb{N}$, $c\in {\cal C} \setminus \{0\}$ and the backward observation $O(\Sigma,D)$, Table 1 reports when $\Sigma$ \emph{entails} a ground streaming atom $\alpha$ (denoted $\Sigma\models \alpha$) or its negation ($\Sigma\models  \mathtt{not} \; \alpha$). If $\Sigma\models\alpha$ ( $\Sigma\models  \mathtt{not} \; \alpha$) we say that $\alpha$ is true (false) at time point $n$.

\begin{table}[t!]

\caption{Entailment of ground streaming literals.}
\begin{tabular}{c c c} 
\toprule
$\alpha$ & $\Sigma\models \alpha$ & $\Sigma\models  \mathtt{not} \; \alpha$ \\
\midrule
$a \;\mathbf{at \;least}\; c \;\mathbf{in} \; \{d_1,\dots,d_m\} $ & $\vert\{A\in O(\Sigma,D): a \in A\}\vert\geq c $&$  \vert\{A\in O(\Sigma,D): a \in A\}\vert < c$\\ 


$a\; \mathbf{always\; in} \;\{d_1,\dots,d_m\}$&$ \forall A\in O(\Sigma,D),  a \in A $ & $\exists A\in O(\Sigma,D): a \not\in A $\\ 
$a \;\mathbf{count}\; c \;\mathbf{in} \; \{d_1,\dots,d_m\}$ & $\vert\{A\in O(\Sigma,D): a \in A\}\vert = c $&$  \vert\{A\in O(\Sigma,D): a \in A\}\vert \neq c$\\

\bottomrule
\end{tabular}

\label{table:entails}

\end{table}

To make a comparison between $\textit{LARS}_{\textit{D}}$ and $\textit{LDSR}$, we defined also for $\textit{LDSR}$ a model-theoretic semantics that can be shown to be equivalent to the operational semantics originally defined. Moreover, besides the concept of streaming model for $\textit{LDSR}$, we defined the concept of answer stream.

Consider an $\textit{LDSR}$ program $P$.
Given a rule $r\in P$, the \emph{ground instantiation} $Gr(r)$ of $r$ denotes the set of rules obtained by applying all possible substitutions $\sigma$ from the variables in $r$ to elements of $\cal{C}$.  In particular, to the counting terms are applied only constants belong to $\cal{C}\cap \mathbb{N^+}$. Similarly, the \emph{ground instantiation} $Gr(P)$ of $P$ is the set $ \bigcup_{r \in P} Gr(r) $.
Given a ground rule $r \in P$, a stream $\Sigma=\langle S_0, \dots, S_n \rangle$ and a stream $\Sigma^{\prime}=\langle S^{\prime}_0, \dots,S^{\prime}_n \rangle$ such that $\Sigma\subseteq \Sigma^{\prime}$, we say that  $\Sigma^{\prime}$ is a model of $r$ for $\Sigma$, denoted $\Sigma^{\prime}\models r$, if $\Sigma^{\prime}\models H(r)$ when $\Sigma^{\prime}\models B(r)$.
We say that $\Sigma^{\prime}$ is a {\em model} of $P$ for $\Sigma$, denoted $\Sigma^{\prime}\models P$, if $\Sigma^{\prime}\models r$ for all rules $r\in Gr(P)$.
Moreover, $\Sigma^{\prime}$ is a {\em minimal model}, if no model $\Omega$ of $P$ exists such that $\Omega \subset \Sigma^{\prime}$ and $t(\Omega)=t(\Sigma')$.
Let  $\Sigma^{\prime}$ be a model of $P$ for $\Sigma$, an atom $a\in S_n'$ is {\em temporary} in  $\Sigma^{\prime}$ if $a \not\in S_n$ and there exists no rule $r\in \mathit{form}_{(1)}(P)$ such that $\Sigma^{\prime}\models B(r)$ and $\Sigma^{\prime}\models H(r)$.
Accordingly, let $T(S_n')$ be the set of all temporary atoms in $S_n'$, the stream $\langle S_0', \dots, S_n' \setminus T(S_n') \rangle$ is called the {\em permanent part} of $\Sigma'$.
Eventually, the {\em reduct} of $P$ w.r.t. $\Sigma^{\prime}$, denoted by $P^{\Sigma^{\prime}}$, consists of the rules $r\in Gr(P)$ such that $\Sigma^{\prime} \models B(r)$.

\begin{definition}
Given an $\textit{LDSR}$ program $P$, a stream $\Sigma=\langle S_0, \dots, S_n \rangle$ and a stream $\Sigma^{\prime}=\langle S^{\prime}_0, \dots,S^{\prime}_n \rangle$ such that $\Sigma\subseteq \Sigma^{\prime}$, $\Sigma^{\prime}$ is called \emph{answer stream} and $S_n'$ \emph{streaming model} of $P$ for $\Sigma$ if: $(1)$
if $n>0$, $\langle S^{\prime}_0\rangle$ is the permanent part of the minimal model $M$ of the reduct $P^{M}$ for the stream $\langle S_0 \rangle$; $(2)$
if $n>0$, for all $i \in 1,\dots,n-1$, $\Sigma^{\prime}_{|i}$ is the permanent part of the minimal model $M$ of the reduct $P^{M}$ for the stream $\langle S^{\prime}_{0},\dots,S^{\prime}_{i-1}, S_i \rangle$; and
$(3)$ $\Sigma^{\prime}$ is a minimal model of the reduct $P^{\Sigma^{\prime}}$ for the stream $\langle S^{\prime}_{0},\dots,S^{\prime}_{n-1}, S_n \rangle$.
\end{definition}

\noindent Note that, differently from LARS, for which the information associated with each time point is entirely derived at the time point of evaluation,  for $\textit{LDSR}$, each time point $t$ in the answer stream is associated with the information derived when the time point $t$ has been evaluated. In other words, the answer stream for $\textit{LDSR}$ is obtained by collecting the results of the previous time points and integrating them with the result of the time point of evaluation.

\section{Framework}\label{sec:framework}
In this section, we present the framework that has been designed for comparing the languages $\textit{LARS}_{\textit{D}}$ and $\textit{LDSR}$.
 The comparison focuses on different parts of the answer stream. In particular,
given an input stream $S=\langle S_0,\dots, S_n\rangle$, when referring to an evaluation time point $t\leq n$, one could compare  the answer streams only at $t$, or in all the time points up to $t$ or also in all time points up to $n$. To this aim, we define three types of streams.
Given $n\in \mathbb{N}$ and $t\in \{0,…,n\}$,
we say that a stream $S=\langle S_0,\dots, S_n\rangle$ is of $type$ $t\mbox{-}atomic$ if $S_i=\emptyset$ for each $i\in \{0,\dots,n\}\setminus \{t\}$; $t\mbox{-}bound$	 if $S_i=\emptyset$ for each $i\in \{t+1,\dots,n\}$;  and $t\mbox{-}\mathit{full}$ if $S_i$ may be nonempty for each $i\in \{0,\dots,n\}$.

Consider a language $L\in \{\textit{LDSR},\textit{LARS}_{\textit{D}}\}$, an input stream $I=\langle I_0,\dots, I_n\rangle$, a set of ground predicate atoms $B\subseteq G_{B}^{\varepsilon} $ 
and a program $P\in L$,  we call $(I,B,P)$ an $L\mbox{-}$tuple.
According to the three types of streams,  for a $L\mbox{-}$tuple  and for each time point $t \in\{0,\dots,n\}$,  we now define three types of output streams for each language $L$.

Given a $\textit{LDSR}\mbox{-}tuple$ $(I,B,P)$ and a time point $t \in \{0,\dots,n\}$, we define:
\begin{itemize}
\item $t\mbox{-}atomic(I,B,P)=\langle O_0,\dots, O_n\rangle$ where $O_i=\emptyset$ for $i\neq t$ and $O_t$ is the streaming model of $P\cup \{b. | b \in B\}$ on $I_{|_t}=\langle I_0,\dots, I_t\rangle$.
\item $t\mbox{-}bound(I,B,P)=\langle O_0,\dots,O_n\rangle$, where $\langle O_0,\dots,O_t\rangle$ is the answer stream of $P\cup \{b. | b \in B\}$ for $I_{|_t}=\langle I_0,\dots,I_t\rangle$ and $O_i=\emptyset$ for $t < i \leq n$.
\item $t\mbox{-}\mathit{full}(I,B,P)=\langle O_0,\dots, O_n\rangle$ where $\langle O_0,\dots,O_t\rangle$ is the answer stream of $P\cup \{b. | b \in B\}$ for $I_{|_t}=\langle I_0,\dots,I_t\rangle$ and $O_i=I_i\cup B$ for $t < i \leq n$.
\end{itemize} 
Analogously, given a $\textit{LARS}_{\textit{D}}\mbox{-}tuple$ $(I,B,P)$ and a time point $t \in \{0,\dots,n\}$, and the answer stream $AS(P,I,t)=\langle A_0,\dots, A_n\rangle$, we define: 
\begin{itemize}
\item $t\mbox{-}atomic(I,B,P)=\langle O_0,\dots, O_n\rangle$ where $O_i=\emptyset$ for $i\neq t$ and $O_t=A_t \cup B$.
\item $t\mbox{-}bound(I,B,P)=\langle O_0,\dots,O_n\rangle$, where $O_i=A_i \cup B$ for $0\leq i \leq t$ and $O_i=\emptyset$ for $t < i \leq n$.
\item $t\mbox{-}\mathit{full}(I,B,P)=\langle O_0,\dots, O_n\rangle$ where $O_i=A_i \cup B$ for $0\leq i \leq n$.
\end{itemize}

\smallskip

We now define when a fragment of a language can be expressed in the other one in our framework. 
In particular, we differentiate expressible fragments from strictly expressible fragments.
Given a stream form  $\phi \in \{atomic,bound,\mathit{full}\}$,and $L_1,L_2\in \{\textit{LDSR}$, $\textit{LARS}_{\textit{D}}\}$ with $L_1\neq L_2$, a 
 fragment $F\subset L_1$ is $\phi\mbox{-}expressible$ via $L_2$ if there exists a mapping $\rho: F \rightarrow L_2$ such that, for each $F$\mbox{-}tuple $(I,B,P)$ and for each time point of evaluation $t \in \{0,\dots,n\}$, it holds that 
$t\mbox{-}\phi(I,B,P)=t\mbox{-}\phi(I,B,\rho(P))_{|_{pred(P\cup I\cup B)}}$; moreover, $F$ is $strictly \; \phi\mbox{-}expressible$ via $L_2$ if $t\mbox{-}\phi(I,B,P)=t\mbox{-}\phi(I,B,\rho(P))$.
Basically, for the non strict expressiveness, a translation into the other language is possible but it can involve the addition of auxiliary predicates, while for the strict one, there is a translation that does not require auxiliary predicates and thus for which the outputs coincide without the need of any filtering.

We are now ready to compare the two languages.
The first result of the comparison is that, without any restrictions, the two languages are incomparable. The following two propositions describe the results. The ideas behind the formal demonstrations, which are instead reported in appendix~\ref{sec:appendixP}, are also introduced.

\begin{proposition}\label{prop:incompOne}
$\textit{LARS}_{\textit{D}}$ is not  $atomic\mbox{-}expressible$ via $\textit{LDSR}$.
\end{proposition}
To see this, consider the simple $\textit{LARS}_{\textit{D}}$ program $P_1$=$\{
@_{T-1}\; a \leftarrow \; @_T\; c.\}$.
This program, that expresses that the presence of an atom $c$ in a time point infers the presence of an atom $a$ in the previous time point, is not expressible in $\textit{LDSR}$ since in its semantic the information associated at every time point are relative only to the information received and inferred up to it. 

\begin{proposition}\label{prop:incompTwo}
$\textit{LDSR}$ is not  $atomic\mbox{-}expressible$ via $\textit{LARS}_{\textit{D}}$.
\end{proposition}
To prove this result, consider the following  $\textit{LDSR}$ program $P_2=\{a(Y) \derives a(X), b(X,Y).\}$ where the predicate $a$ belongs to the input predicates ${\cal P}_{S}^{\varepsilon}$. The program $P_2$ is not expressible in $\textit{LARS}_{\textit{D}}$ since its semantics avoids to infer ground atoms over input predicates.

Since $t\mbox{-}atomic(I,B,P) \neq t\mbox{-}atomic(I,B,\rho(P))$ implies $t\mbox{-}\phi(I,B,P)=t\mbox{-}\phi(I,B,\rho(P))$ holds for $\phi \in \{bound,\mathit{full}\}$, Proposition \ref{prop:incompOne} and \ref{prop:incompTwo} imply the following result.

\begin{theorem}
$\textit{LARS}_{\textit{D}}$ and $\textit{LDSR}$ are incomparable under each of the three stream forms.
\end{theorem}

\begin{figure}[t!] 
\centering
  \subfloat[\textit{LARS} fragments]{\includegraphics[width=6cm]{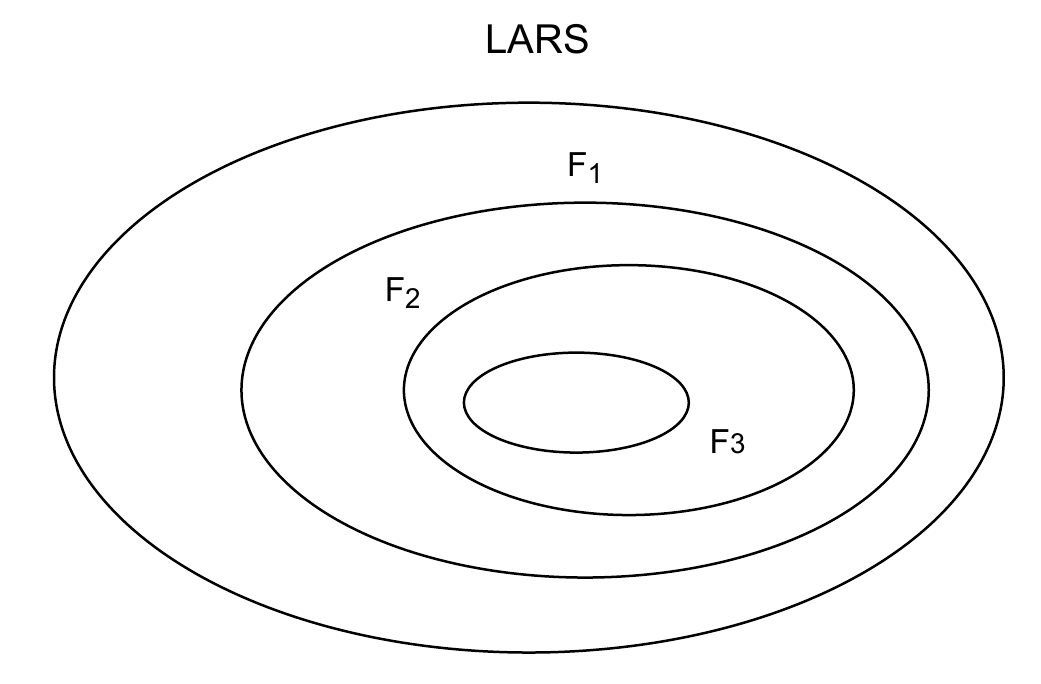}\label{fig1}}
  \hfill
  \subfloat[\textit{LDSR} fragments]{\includegraphics[width=6cm]{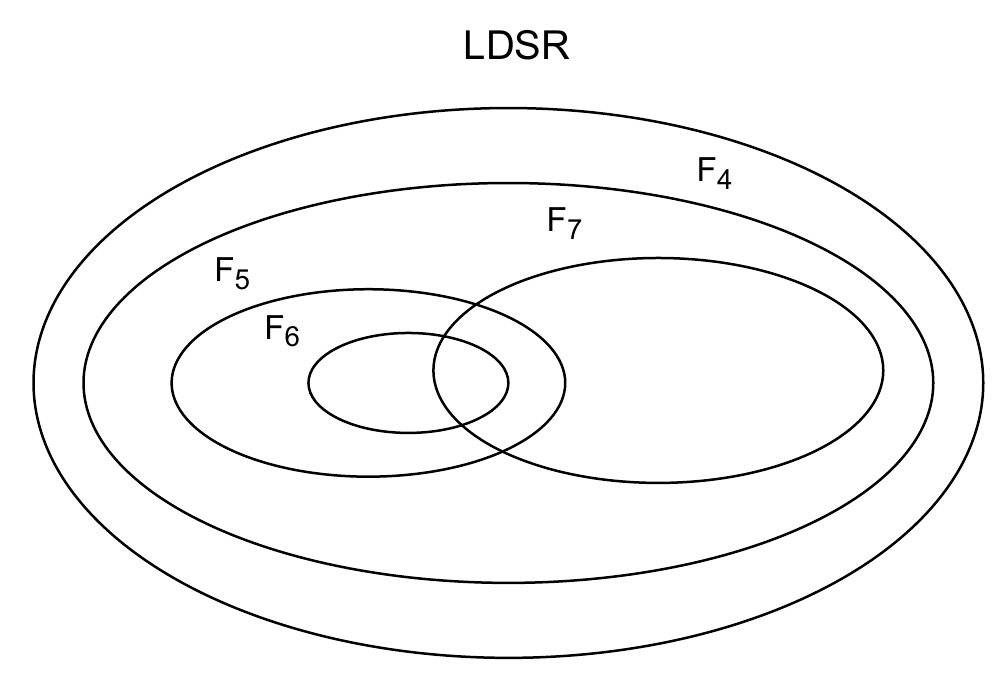}\label{fig2}}
   \caption{Fragments and their relations}\label{fig:fragments}
\end{figure}

Given the incomparability of the languages, we introduced some restrictions to identify fragments of a language expressible in the other. Up to now, we identified seven fragments that are described in detail in sections \ref{sec:LARS to LDSR} and \ref{sec:LDSR to LARS}. Here, we briefly discuss their relations and expressiveness (see Fig.~\ref{fig:fragments} and Tables \ref{Table:LARStoLDSR} and \ref{Table:LDSRtoLARS}). In particular, Table~\ref{Table:LARStoLDSR}  presents the fragments of $\textit{LARS}_{\textit{D}}$ $F_1$, $F_2$ and $F_3$. All of them are strictly expressible via $\textit{LDSR}$. $F_1$ is the largest identified fragment and it is atomic\mbox{-}expressible; $F_2$ is obtained from $F_1$ by imposing some restriction and it is bound\mbox{-}expressible, while $F_3$ is obtained by further restricting $F_2$, and allows for achieving full\mbox{-}expressivity. 
\begin{table}[t!]
\parbox{.45\linewidth}{
\centering
\caption{$\textit{LARS}_{\textit{D}}$ to $\textit{LDSR}$}
\label{Table:LARStoLDSR}
\begin{tabular}{c c c } 
\toprule
$\phi$ & strictly & not strictly \\
\midrule
$atomic$ & $F_1$ & $F_1$ \\ 
$bound$ & $F_2$ & $F_2$  \\ 
$\mathit{full}$ & $F_3$ & $F_3$  \\ 
\bottomrule
\end{tabular}
}
\hfill
\parbox{.45\linewidth}{
\centering
\caption{$\textit{LDSR}$ to $\textit{LARS}_{\textit{D}}$}
\label{Table:LDSRtoLARS}
\begin{tabular}{c c c } 
\toprule
$\phi$ & strictly & not strictly \\
\midrule
$atomic$ & $F_6 \cup F_7$ & $F_4$ \\ 
$bound$ & $F_6 \cup F_7$ & $F_4$  \\ 
$\mathit{full}$ & $F_6$ & $F_5$  \\
\bottomrule
\end{tabular}
}
\end{table}

As for $\textit{LDSR}$, Table~\ref{Table:LDSRtoLARS} summarizes its identified fragments $F_4$,$F_5$,$F_6$ and $F_7$. The largest fragment is $F_4$, which is bound\mbox{-}expressible via $\textit{LARS}_{\textit{D}}$; the fragment $F_5$, that restricts $F_4$, allows full expressiveness; the other two fragments allows for strictly expressiveness: $F_6$ further restricts $F_5$, and it is full\mbox{-}expressible, while $F_7$ is obtained by imposing restrictions on $F_4$ and is bound\mbox{-}expressible.

\subsection{$\textit{LARS}_{\textit{D}}$ to $\textit{LDSR}$}\label{sec:LARS to LDSR}
Here we present the identified fragments of $\textit{LARS}_{\textit{D}}$. We consider two types of rules:

\begin{equation*}
\mbox{$(I)$} \ \ \ \square( a \leftarrow \beta_1\wedge\dots\wedge\beta_b)
\ \ \ \ \ \ \ \  \mbox{and}
\ \ \ \ \ \ \ \ 
\mbox{$(\mathit{II})$} \ \ \ a \leftarrow \beta_1,\dots,\beta_b
\end{equation*}

\noindent where $a$ is an intensional predicate atom and $\beta_i\in \{\boxplus^m \lozenge p, \boxplus^m \square p, p, \boxplus^0 @_T \top \wedge @_{T-K}p, \\ \neg \boxplus^m\lozenge p, \neg \boxplus^m \square p, \neg p,  \neg (\boxplus^0 @_T \top \wedge @_{T-K}p) | p$ is a predicate atom, $T\in \mathcal{U}$ and $m\in \mathbb{N} \}$ for $i\in \{1,\dots,b\}$.
 For a rule of type $(I)$, we denote as  $cons(r)$ the consequent $\{a\}$ of the implication, and with $prem(r)$ the set of formulas $\{\beta_1,\dots,\beta_b\}$ in the premise; moreover, for a rule of type $(\mathit{II})$, we denote with $head(r)$ the atom $\{a\}$ and with $body(r)$ the set of formulas $\{\beta_1,\dots,\beta_b\}$.  
Let $P$ be a $\textit{LARS}_{\textit{D}}$ program, we denote with $type_I(P)$ the set of rules of $P$ of type $(I)$ and with $type_{\mathit{II}}(P)$ the set of rules of $P$ of type $(\mathit{II})$. \\
We say that a predicate  $p$ is $\mathit{marked}$ if there are two rules $r\in type_{I}(P)$ and $r'\in type_{\mathit{II}}(P)$ with $p=pred(cons(r))$, $h\in pred(prem(r))$ and $h= pred(head(r'))$. The set of marked predicates of a program $P$ is denoted with $M(P)$. 

Consider a $\textit{LARS}_{\textit{D}}$ program $P$ containing rules of type $(I)$ and $(\mathit{II})$ only. Let $P_1 = type_{I}(P)$ and $P_2 = type_{\mathit{II}}(P)$. We define the graph $G(P)=\langle N, A \rangle$, where: $(1)$ $N=(\cup_{r \in P_1}pred(cons(r)))$ $\cup$ $(\cup_{r \in P_2}pred(head(r)))$; $(2)$ 
$(q,p,\quot{$+$})\in A$ if there exists a rule $r\in P_1$ with $pred(cons(r))=p$ and $q\in pred(prem(r))$ occurring in a formula without negation or if there exists a rule $r'\in P_2$ with $pred(head(r'))=p$ and $q\in pred(body(r'))$ occurring in a formula without negation; and $(3)$ $(q,p,\quot{$-$})\in A$ if there exists a rule $r\in P_1$ with $pred(cons(r))=p$ and $q\in prem(r)$ occurs in a formula with negation or  if there exists a rule $r'\in P_2$ with $pred(head(r'))=p$ and $q\in pred(body(r'))$ occurring in a formula with negation.

\begin{definition}
Fragment $F_1$ of $\textit{LARS}_{\textit{D}}$ collects all the programs $P$ that meet the next conditions:
$(i)$ $P=type_{I}(P)\cup type_{\mathit{II}}(P)$; 
$(ii)$ $\cup_{r \in type_{I}(P)}pred(prem(r))\cap M(P)=\emptyset$;  $(iii)$ $\cup_{r \in type_{\mathit{II}}(P)}pred(body(r))\cap M(P)=\emptyset$; 
$(iv)$ no cycle in $G(P)$ contains an arc labeled with \quot{-}.
\end{definition}
Roughly, $F_1$ contains only programs that are stratified w.r.t negation, featuring only rules of types (I) and (II), where no marked predicate appears in premises and in bodies.

\begin{proposition}
$F_1$ is strictly atomic\mbox{-}expressible via $\textit{LDSR}$.
\end{proposition}
Indeed, it can be shown that there is a mapping  $\rho_1:F_1\rightarrow \textit{LDSR}$ 
such that for each $F_1\mbox{-}tuple$ $(I,B,P)$ and for each time point of evaluation $t \in \{0,\dots,n\}$, it holds that $t\mbox{-}atomic(I,B,P)=t\mbox{-}atomic(I,B,\rho_1(P))$. 
In particular, given a program $P\in F_1$, the $\textit{LDSR}$ program $\rho_1(P)$ is obtained by replacing: 
\begin{itemize}
\item[-]each rule $\square (a \leftarrow \beta_1,\dots,\beta_m$) of type $(I)$ with the $\textit{LDSR}$ rule $\;a \derives\; f(\beta_1),\dots,f(\beta_m)$ of form $(1)$
\item[-]each rule $a \leftarrow \beta_1,\dots,\beta_m$ of type $(II)$ with the $\textit{LDSR}$ rule $\mathbf{\#temp} \; a \derives\; f(\beta_1),\dots,f(\beta_m)$ of form $(2)$
\end{itemize}
where $f$ associates each $\textit{LARS}_{\textit{D}}$ formula in the $F_1$ fragment with a $\textit{LDSR}$ streaming atom as reported below:
\begin{itemize}
\item[-]$f(\boxplus^m \lozenge p)=p \;\mathbf{in}\;[m]$.
\item[-]$f(\boxplus^m \square p)=p \;\mathbf{always \; in}\;[m]$.
\item[-]$f(p)=p$.

\item[-]$f(\boxplus^0 @_T \top \wedge @_{T-K}p)=p \;\mathbf{in} \; \{k\}$.
\item[-]$f(\neg \beta)=\mathtt{not}\; f(\beta)$ where $\beta$ is a formula.
\end{itemize}
Intuitively, the idea of the mapping $\rho_1$ is that $\textit{LARS}_{\textit{D}}$ rules of type $(I)$ that must be evaluated at each time point are associated with $LDRS$ rules of form (1) and $\textit{LARS}_{\textit{D}}$ rules of type $\mathit{(II)}$ that infer information only at the evaluation time point $t$ are associate with $\textit{LDSR}$ rules of form (2) which are evaluated at each time point but the derivations of the time points preceding $t$ have been forgotten via the $\mathbf{\#temp}$ operator.
Moreover, we impose the restrictions $(ii)$ and $(iii)$ for defining $F_1$ in order to achieve atomic-expressiveness; indeed, they ensure that, if in $\textit{LDSR}$ a permanent information is derived relying on a temporary information, this is not used to derive other information.

Now, we define the fragment $F_2$ that imposes an additional restriction w.r.t to $F_1$, and the fragment $F_3$ that further restricts $F_2$.
\begin{definition}
The fragment $F_2$ of $\textit{LARS}_{\textit{D}}$ is the subset of the programs of $F_1$ that meet the condition 
$(\bigcup_{r \in type_{\mathit{II}}(P)}pred(head(r)))\cap ((\bigcup_{r \in type_{I}(P)}pred(prem(r))=\emptyset$.
\end{definition}
\begin{proposition}\label{bound}
$F_2$ is strictly bound\mbox{-}expressible via $\textit{LDSR}$.
\end{proposition}
It can be shown that for the mapping $\rho_2={\rho_1}_{\vert_ {F_2}}$, it holds that for each $F_2\mbox{-}tuple$ $(I,B,P)$ and for each time point of evaluation $t \in \{0,\dots,n\}$,  $t\mbox{-}bound(I,B,P)=t\mbox{-}bound(I,B,\rho_2(P))$.
Basically, the additional condition for fragment $F_2$ avoids that a temporary information associated to a time point can generate a permanent information.

We now show an example of a program belonging to fragment $F_2$  and its image with respect to the function $\rho_2$. The example is taken from one of the tasks of the ``model and
solve'' Stream Reasoning Hackathon 2021~\cite{DBLP:conf/esws/SchneiderALDP22}. 
The task concerns urban traffic management. Traffic is observed from a top-down, third-person perspective, and vehicle movement flows in a given road network coded as Datalog facts are considered. 
We want to identify the vehicles that appear or disappear in the network. It is then necessary to note vehicles that were absent at the previous time point and are now present and vice versa. 
A rule of type $(I)$ is used to evaluate the presence of vehicles at the current and previous time points; then the obtained information is used in a rule of type $(\mathit{II})$ that detects the appearance or disappearance of a vehicle at the evaluation time point.
This task can be modelled via the following program $P$ in $F_2$:
\begin{multline*}
\ \ \ \ \ \ \ \ \ \ \ \ \ \ \ \ \  \square(inNetwork(Veh)\leftarrow onLane(Veh,X,Y)).  \\
 \ \ \ \ \ \ \ \ \ \ \ \ \ \ \ \ \ appears(Veh) \leftarrow onLane(Veh,X,Y), \neg \boxplus^0 @_T \top \wedge @_{T-1}inNetwork(Veh). \\
disappears(Veh) \leftarrow \boxplus^0 @_T \top \wedge @_{T-1}inNetwork(Veh), \neg inNetwork(Veh).
\end{multline*}
The corresponding \textit{LDSR} program, image of the $\rho_2$ function, is:
\begin{multline*}
\ \ \ \ \ \ \ \ \ \ \ \ \ \ inNetwork(Veh)\derives onLane(Veh,X,Y).\\
\ \ \ \mathbf{\#temp} \;  appears(Veh) \derives onLane(Veh,X,Y), \mathtt{not}\   inNetwork(Veh) \;\mathbf{in} \; \{1\}. \\
\mathbf{\#temp} \;  disappears(Veh) \derives onLane(Veh,X,Y) \;\mathbf{in} \; \{1\}, \mathtt{not} \ inNetwork(Veh).
\end{multline*}

\begin{definition}
The fragment $F_3$ of $\textit{LARS}_{\textit{D}}$ is the subset of the programs of $F_2$ that meet the condition $P=type_{\mathit{II}}(P)$.
\end{definition}
\begin{proposition}
$F_3$ is strictly full\mbox{-}expressible via $\textit{LDSR}$.
\end{proposition}
Similarly to $F_2$, it can be shown that, considering the mapping  $\rho_3={\rho_1}_{\vert_ {F_3}}$, it holds that that for each $F_3\mbox{-}tuple$ $(I,B,P)$ and for each time point of evaluation $t \in \{0,\dots,n\}$,  $t\mbox{-}\mathit{full}(I,B,P)=t\mbox{-}\mathit{full}(I,B,\rho_3(P))$.
Intuitively, since in $\textit{LDSR}$ the evaluation of a program with respect to a time point $t$ can not add information in the output stream at time points that are subsequent to $t$, to achieve full expressiveness, we impose the restriction to rules of type type $\mathit{(II)}$ in $\textit{LARS}_{\textit{D}}$ as these are evaluated only at $t$ and do not change the output in the subsequent time points. 

\subsection{$\textit{LDSR}$ to $\textit{LARS}_{\textit{D}}$}\label{sec:LDSR to LARS}
Here we present the identified fragments of $\textit{LDSR}$ along with their expressiveness. 
While, for the fragments of $\textit{LARS}_{\textit{D}}$
we obtained strict expressiveness and each rule in $\textit{LARS}_{\textit{D}}$  has been translated into exactly one rule in $\textit{LDSR}$, for the fragments of $\textit{LDSR}$, the translation, in general, needs auxiliary atoms and additional rules to simulate the behavior of rules of the form (2) and of some streaming and aggregates atoms. 

The largest identified fragment is $F_4$ that is defined as follows.

\begin {definition}
The fragment $F_4$ is the subset of the $\textit{LDSR}$ programs $P$ that meet the condition $\; \bigcup_{r\in P}pred(H(r))\subset {\cal P}^I$.

\end{definition}
Basically, $F_4$ is obtained from $\textit{LDSR}$ by simply imposing that no extensional predicate appears in the head of a rule.
\begin{proposition}
$F_4$ is bound\mbox{-}expressible via $\textit{LARS}_{\textit{D}}$.
\end{proposition}
It can be shown that there is a mapping  $\rho_4:F_4\rightarrow \textit{LARS}_{\textit{D}}$ 
such that for each $F_4\mbox{-}tuple$ $(I,B,P)$ and for each $t \in \{0,\dots,n\}$,  $t\mbox{-}bound(I,B,P)=t\mbox{-}bound(I,B,\rho_4(P))_{|_{pred(P\cup I\cup B)}}$. 
First, we note that, in general, each streaming atom in $F_4$ has to properly translated into a $\textit{LARS}_{\textit{D}}$ formula; moreover  a special rewriting, requiring additional rules, has to be performed for streaming atoms of the form $a  \;\mathbf{count}\; v \;\mathbf{in} \; \{ d_1,\dots,d_m\}$, where $v$ is a variable in $\cal{C}$ and for all the aggregate atoms.
Thus, without going into details, the mapping $\rho_4$ relies on a function $g$ that associates each streaming atom (but those of form $a \;\mathbf{count}\; v \;\mathbf{in} \; \{ d_1,\dots,d_m\}$) with a $\textit{LARS}_{\textit{D}}$ formula that expresses the condition that must be satisfied in the stream for the streaming atom to be true; moreover, if $\alpha$ is an aggregate atom or a streaming atom of the form $a  \;\mathbf{count}\; v \;\mathbf{in} \; \{ d_1,\dots,d_m\}$, $g$ associates it with a formula containing auxiliary atoms defined via an additional set of rules $C_{\alpha}$ that are needed to simulate its semantics. For the sake of the presentation, the function $g$ and the set of additional rules $C_{\alpha}$ are reported in appendix~\ref{sec:appendixA}

Furthermore, given a program $P\in F_4$, the mapping $\rho_4$
has to replace each rule $r$ in $P$ with one or more $\textit{LARS}_{\textit{D}}$ rules.
In sum, the program $\rho_4(P)$ is obtained by:
\begin{itemize}
\item[-]replacing each rule $\;a \derives\; \beta_1,\dots,\beta_m$ of form (1) with the $\textit{LARS}_{\textit{D}}$ rule \\  $\square (a \leftarrow \boxplus^0 @_T \top\wedge g(\beta_1)\wedge\dots\wedge g(\beta_m))$.
\item[-]for each rule $\; \mathbf{\#temp} \; a \derives\; \beta_1,\dots,\beta_m$ of form (2), 
\begin{itemize}
\item[-] replacing it with the $\textit{LARS}_{\textit{D}}$ rule $a \leftarrow \boxplus^0 @_T \top, g(\beta_1),\dots,g(\beta_m)$
\item[-]  adding the rule $\square (atemp \leftarrow \boxplus^0 @_T \top\wedge g(\beta_1)\wedge\dots\wedge g(\beta_m))$.
\end{itemize}
\item[-] adding for each streaming atom $\alpha$ of the form  $a  \;\mathbf{count}\; v \;\mathbf{in} \; \{ d_1,\dots,d_m\}$, where $v$ is a variable in $\cal{C}$ or aggregates atom, a set of rules $C_{\alpha}$. 
 

\end{itemize}
Basically, the mapping $\rho_4$ manages the interaction between the two forms of rules, simulating, at an evaluation time point $t$, the temporary derivations obtained in the previous time points and evaluating their effect on the permanent derivations that will be part of the output. This is mainly obtained by creating and handling a copy of each rule of the form (2) where the suffix ``temp'' is added to the head predicate.
Moreover, since the streaming atoms in $\textit{LDSR}$ are evaluated according to the backward observation, we need to identify the reference time point in which the $LARS$ formulas representing the conditions expressed by the streaming atoms have to checked. To do this, we use the formula $\boxplus^0 @_T \top$ that evaluates the tautology within a window of size 0 and thus, it holds in the rule instance where the variable $T$ corresponds to the reference time point.

We are now ready to define the fragment $F_5$ that is obtained from $F_4$ by avoiding rules of form (1).

\begin {definition}
The fragment $F_5$ of $\textit{LDSR}$ is the subset of the programs of $F_4$ that meet the condition $P= \mathit{form}_{2}(P)$.
\end{definition}

\begin{proposition}
$F_5$ is $\mathit{full}\mbox{-}expressible$ via $\textit{LARS}_{\textit{D}}$.
\end{proposition}
 To see this, consider, the mapping $\rho_5: F_5 \rightarrow \textit{LARS}_{\textit{D}}$ such that, for each program $P\in F_5$, $\rho_5(P)$ is obtained by:  
\begin{itemize}
\item[-]replacing each rule $\; \mathbf{\#temp} \; a \derives\; \beta_1,\dots,\beta_m$ of form (2) , with the $\textit{LARS}_{\textit{D}}$ rule \\ $a \leftarrow \boxplus^0 @_T \top, g'(\beta_1),\dots,g'(\beta_m)$.
\item[-]adding for each streaming atom $\alpha$ of the form  $a  \;\mathbf{count}\; v \;\mathbf{in} \; \{ d_1,\dots,d_m\}$, where $v$ is a variable in $\cal{C}$ or aggregates atom, a set of rules $C_{\alpha}$. 
\end{itemize}
It can be shown that $\rho_5$ is  
such that for each $F_5\mbox{-}tuple$ $(I,B,P)$ and for each $t \in \{0,\dots,n\}$,  $t\mbox{-}\mathit{full}(I,B,P)=t\mbox{-}\mathit{full}(I,B,\rho_5(P))_{|_{pred(P\cup I\cup B)}}$.
Roughly, similarly to $\rho_4$, $\rho_5$ relies on a function $g'$ for rewriting body atoms and adds auxiliary rules for handling aggregate atoms or a streaming atoms of the form $a  \;\mathbf{count}\; v \;\mathbf{in} \; \{ d_1,\dots,d_m\}$ (more details on this are reported in Appendix in\ref{sec:appendixB}); however the translation for the fragment $F_5$ is simpler than  the one for $F_4$, as it is sufficient to associate each rule of form (2) with a $\textit{LARS}_{\textit{D}}$ rule of type (II). Indeed, since rules of form (1) are not allowed, there is no need to consider the information that can be derived in a permanent way through them. 
 The condition defining the fragment $F_5$ ensures the full expressiveness: since a program in this fragment features only rules of form (2), and its translation only rules of type (II), their evaluation at each time point $t$ can derive information only at $t$, while leaving unchanged the output in the other time points.

The fragment $F_6$ restrict $F_5$ to reach strict full expressiveness.

\begin {definition}
The fragment $F_6$ of $\textit{LDSR}$ is the subset of the programs of $F_5$ in which streaming atoms of the form $a  \;\mathbf{count}\; v \;\mathbf{in} \; \{ d_1,\dots,d_m\}$ where $v\in V$ and aggregate atoms are not allowed in rule bodies.
\end{definition}

\begin{proposition}
$F_6$ is strictly full\mbox{-}expressible via $\textit{LARS}_{\textit{D}}$.
\end{proposition}
It can be shown that for the mapping $\rho_6={\rho_5}_{\vert_ {F_6}}$, it holds that for each $F_6\mbox{-}tuple$ $(I,B,P)$ and for each $t \in \{0,\dots,n\}$,  $t\mbox{-}\mathit{full}(I,B,P)=t\mbox{-}\mathit{full}(I,B,\rho_6(P))$. Since no atoms involving the addition of auxiliary predicates are considered, $F_6$ is strictly expressible.

The last considered fragment $F_7$ still features strict expressiveness, but of bound type, as, differently from $F_6$, it allows also rules of form (1) to some extent.

\begin {definition}
The fragment $F_7$ of $\textit{LDSR}$ is the subset of the programs of $F_4$ that meet the following conditions: $(i)$ $(\cup_{r\in \mathit{form}_2(P)}(pred(B(r)))\cap (\cup_{r\in \mathit{form_2(P)}}(pred(H(r)))=\emptyset $
%
$(ii)$  streaming atoms of the form $a  \;\mathbf{count}\; v \;\mathbf{in} \; \{ d_1,\dots,d_m\}$ where $v\in V$ and aggregate atoms are not allowed in rule bodies.
%
\end{definition}

\begin{proposition}
$F_7$ is strictly bound\mbox{-}expressible via $\textit{LARS}_{\textit{D}}$.
\end{proposition}
It can be shown that there is a mapping  $\rho_7:F_7\rightarrow \textit{LARS}_{\textit{D}}$ 
such that for each $F_7\mbox{-}tuple$ $(I,B,P)$ and for each $t \in \{0,\dots,n\}$,  $t\mbox{-}bound(I,B,P)=t\mbox{-}bound(I,B,\rho_7(P))$. 
To see this, consider, the mapping $\rho_7: F_7 \rightarrow \textit{LARS}_{\textit{D}}$ such that, for each program $P\in F_7$, $\rho_7(P)$ is obtained by:  
\begin{itemize}
\item[-]replacing each rule $\; \mathbf{\#temp} \; a \derives\; \beta_1,\dots,\beta_m$ of form (2) with the $\textit{LARS}_{\textit{D}}$ rule \\ $a \leftarrow \boxplus^0 @_T \top, g'(\beta_1),\dots,g'(\beta_m)$.
\item[-]replacing each rule $\;a \derives\; \beta_1,\dots,\beta_m$ 
of form (1) with the $\textit{LARS}_{\textit{D}}$ rule \\
$\square (a \leftarrow \boxplus^0 @_T \top\wedge g''(\beta_1)\wedge\dots\wedge g''(\beta_m))$.
\end{itemize}

\noindent The mapping relies on the same function $g'$ as $\rho_5$ for the rules of form (2).
In addition, for the rule of form (1), a different function $g''$ is used to associate the body streaming atoms with $LARS$ formulas based on the following definition.

\begin{definition}
 Given an atom $a(t_1,\dots,t_n)$ and a $\textit{LDSR}$ rule $r$ with $H(r)=a(t_1',\dots,t_n')$ we call definition of $a(t_1,\dots,t_n)$ in $r$, denoted with $d_r(a(t_1,\dots,t_n))$, the conjunction $\bigwedge_{\{\beta\in B(r)\}}\beta \bigwedge_{1 \leq i\leq n}t_i=t_i'$. 
Given an \textit{LDSR} program $P$ the definition of $a(t_1,\dots,t_n)$ in $P$ is $d_P(a(t_1,\dots,t_n))=a(t_1,\dots,t_n)\vee (\bigvee_{\{r\in P| pred(H(r))= a\}} d_r(a(t_1,\dots,t_n)))$. 
\end{definition}
This definition identifies, for each predicate $a$ in the head of a rule of form (2), a $LARS$ formula relying only on permanent information that can be used in the $LARS$ translation in place of $a$.  
Further details on the translation and the $g''$ functions are reported in Appendix~\ref{sec:appendixC}. We note here that the strict expressiveness of this fragment is obtained since the translation of the allowed streaming atoms does not require the use of additional atoms, and condition $(ii)$ simplifies the rewriting of the rules of form $(2)$ w.r.t what fragment $F_4$. Indeed, in this case, it is not necessary to add the rules used by $\rho_4$ such as $\square (atemp \leftarrow \boxplus^0 @_T \top\wedge g(\beta_1)\wedge\dots\wedge g(\beta_m))$ that required additional auxiliary atoms. 

Consider, for example, the $\textit{LDSR}$ program $P'$ that could be used for monitoring irregularity in a subway station monitoring system. Three minutes are expected to elapse between the arrival of one train and the next, so the program records an irregularity when one train passes and another has already passed in one of the previous two minutes:
\[P'=\{irregular \ \derives train\_pass, train\_pass\;\mathbf{at \;least}\; 1 \;\mathbf{in} \; \{1,2\}.\}\] The program belongs to the $F_7$ fragment, and the corresponding $\textit{LARS}_{\textit{D}}$ program with respect to the $\rho_7$ function is as follows:
\begin{multline*}
\rho_7(P')=\{\square (irregular \leftarrow \boxplus^0 @_T \top\wedge (@_{T_1}train\_pass\wedge T_1=T-0) \\ \wedge(@_{T_2}train\_pass\wedge ((T_2=T-1)\vee(T_2=T-2))). \}
\end{multline*}






\section{Conclusion}
This work presents a formal comparison about the relative expressiveness of the two languages $\textit{LDSR}$ and \textit{LARS}. The main contribution of the work is twofold: $(i)$ we propose a suitable framework to compare the two languages, which exhibit different syntax and semantics. and $(ii)$ for each language, we identify a number of fragments that can be expressed by the other one, showing possible rewritings. 
In order to compare the semantics  of the two languages, we first provided an alternative equivalent model-theoretic definition of the semantics of $\textit{LDSR}$, instead of the operational one originally provided. Moreover, 
we defined the concept of answer stream also for $\textit{LDSR}$, as an extension of the streaming model.
The framework allowed us for focusing the comparison on different forms of the output stream (atomic, bound, full) and  on the nature of the rewriting that could forbid or admit the addition of auxiliary predicates (strict or not strict expressiveness, respectively).
For each given form of output and  type of rewriting, we studied how to build fragments of a language that could meet the desired expressiveness. To do this, we  considered the semantics behind each construct or combination of constructs that can occur in the rules and the effect of interactions between the different rules. The fragments $F_1,...,F_7$ are the largest we identified so far, but, of course, these could be further enlarged and new ones could be possibly found; this will be the subject of future works.

\section{Acknowledgments}
 This work has been partially supported by the project “MAP4ID - Multipurpose Analytics Platform 4 Industrial Data”, N. F/190138/01-03/X44 and by the Italian MIUR Ministry and the Presidency of the
Council of Ministers under the project \quot{Declarative Reasoning over
Streams} under the \quot{PRIN} 2017 call (CUP $H24I17000080001$, project
2017M9C25L\_001).

\bibliographystyle{splncs.bst}
\bibliography{main.bib}

\begin{thebibliography}{10}

\bibitem{DBLP:journals/datasci/DellAglioVHB17}
Dell'Aglio, D., Valle, E.D., van Harmelen, F., Bernstein, A.:
\newblock Stream reasoning: {A} survey and outlook.
\newblock Data Sci. \textbf{1}(1-2) (2017)  59--83

\bibitem{DBLP:journals/ijsc/BarbieriBCVG10}
Barbieri, D.F., Braga, D., Ceri, S., Valle, E.D., Grossniklaus, M.:
\newblock {C-SPARQL:} a continuous query language for {RDF} data streams.
\newblock Int. J. Semantic Comput. \textbf{4}(1) (2010)  3--25

\bibitem{hoeksema2011high}
Hoeksema, J., Kotoulas, S.:
\newblock High-performance distributed stream reasoning using s4.
\newblock In: Ordring Workshop at ISWC. (2011)

\bibitem{DBLP:conf/lpnmr/PhamAM19}
Pham, T., Ali, M.I., Mileo, A.:
\newblock {C-ASP:} continuous asp-based reasoning over {RDF} streams.
\newblock In: {LPNMR}. Volume 11481 of LNCS., Springer (2019)  45--50

\bibitem{DBLP:conf/semweb/PhuocDPH11}
Phuoc, D.L., Dao{-}Tran, M., Parreira, J.X., Hauswirth, M.:
\newblock A native and adaptive approach for unified processing of linked
  streams and linked data.
\newblock In: International Semantic Web Conference {(1)}. Volume 7031 of
  LNCS., Springer (2011)  370--388

\bibitem{DBLP:journals/cacm/BrewkaET11}
Brewka, G., Eiter, T., Truszczynski, M.:
\newblock Answer set programming at a glance.
\newblock Communications of the {ACM} \textbf{54}(12) (2011)  92--103

\bibitem{DBLP:conf/ijcai/GebserLMPRS18}
Gebser, M., Leone, N., Maratea, M., Perri, S., Ricca, F., Schaub, T.:
\newblock Evaluation techniques and systems for answer set programming: a
  survey.
\newblock In: {IJCAI}, ijcai.org (2018)  5450--5456

\bibitem{beck2018lars}
Beck, H., Dao{-}Tran, M., Eiter, T.:
\newblock {LARS:} {A} logic-based framework for analytic reasoning over
  streams.
\newblock Artif. Intell. \textbf{261} (2018)  16--70

\bibitem{bazoobandi2017expressive}
Bazoobandi, H.R., Beck, H., Urbani, J.:
\newblock Expressive stream reasoning with laser.
\newblock In: International Semantic Web Conference {(1)}. Volume 10587 of
  LNCS., Springer (2017)  87--103

\bibitem{beck2018ticker}
Beck, H., Dao{-}Tran, M., Eiter, T., Folie, C.:
\newblock Stream reasoning with {LARS}.
\newblock K{\"{u}}nstliche Intell. \textbf{32}(2-3) (2018)  193--195

\bibitem{eiter2019distributed}
Eiter, T., Ogris, P., Schekotihin, K.:
\newblock A distributed approach to {LARS} stream reasoning (system paper).
\newblock TPLP \textbf{19}(5-6) (2019)  974--989

\bibitem{DBLP:conf/bigdataconf/RenCNX18}
Ren, X., Cur{\'{e}}, O., Naacke, H., Xiao, G.:
\newblock Bigsr: real-time expressive {RDF} stream reasoning on modern big data
  platforms.
\newblock In: {IEEE} BigData, {IEEE} (2018)  811--820

\bibitem{DBLP:conf/rr/MileoAPH13}
Mileo, A., Abdelrahman, A., Policarpio, S., Hauswirth, M.:
\newblock Streamrule: {A} nonmonotonic stream reasoning system for the semantic
  web.
\newblock In: {RR}. Volume 7994 of LNCS., Springer (2013)  247--252

\bibitem{DBLP:conf/ai/DoLL11}
Do, T.M., Loke, S.W., Liu, F.:
\newblock Answer set programming for stream reasoning.
\newblock In: Canadian Conference on {AI}. Volume 6657 of LNCS., Springer
  (2011)  104--109

\bibitem{DBLP:conf/lpnmr/GebserGKS11}
Gebser, M., Grote, T., Kaminski, R., Schaub, T.:
\newblock Reactive answer set programming.
\newblock In: {LPNMR}. Volume 6645 of LNCS., Springer (2011)  54--66

\bibitem{DBLP:journals/tplp/CalimeriMMMPZ21}
Calimeri, F., Manna, M., Mastria, E., Morelli, M.C., Perri, S., Zangari, J.:
\newblock I-dlv-sr: {A} stream reasoning system based on {I-DLV}.
\newblock Theory Pract. Log. Program. \textbf{21}(5) (2021)  610--628

\bibitem{carbone2015apache}
Carbone, P., Katsifodimos, A., Ewen, S., Markl, V., Haridi, S., Tzoumas, K.:
\newblock Apache flink{\texttrademark}: Stream and batch processing in a single
  engine.
\newblock {IEEE} Data Eng. Bull. \textbf{38}(4) (2015)  28--38

\bibitem{DBLP:journals/tplp/IanniPZ20}
Ianni, G., Pacenza, F., Zangari, J.:
\newblock Incremental maintenance of overgrounded logic programs with tailored
  simplifications.
\newblock TPLP \textbf{20}(5) (2020)  719--734

\bibitem{DBLP:journals/tplp/CalimeriFGIKKLM20}
Calimeri, F., Faber, W., Gebser, M., Ianni, G., Kaminski, R., Krennwallner, T.,
  Leone, N., Maratea, M., Ricca, F., Schaub, T.:
\newblock Asp-core-2 input language format.
\newblock TPLP \textbf{20}(2) (2020)  294--309

\bibitem{DBLP:conf/esws/SchneiderALDP22}
Schneider, P., Alvarez{-}Coello, D., Le{-}Tuan, A., Duc, M.N., Phuoc, D.L.:
\newblock Stream reasoning playground.
\newblock In Groth, P., Vidal, M., Suchanek, F.M., Szekely, P.A., Kapanipathi,
  P., Pesquita, C., Skaf{-}Molli, H., Tamper, M., eds.: The Semantic Web - 19th
  International Conference, {ESWC} 2022, Hersonissos, Crete, Greece, May 29 -
  June 2, 2022, Proceedings. Volume 13261 of Lecture Notes in Computer
  Science., Springer (2022)  406--424

\end{thebibliography}

\newpage

\appendix
\section{Details on the translation from $\textit{LDSR}$ to $\textit{LARS}_{\textit{D}}$}\label{sec:appendix0}
We introduce a function $\sigma$ that associates any streaming atoms with a $LARS$ formula. Function $\sigma$ will be used to present $g$,$g'$ and $g''$ and works as follows:

\begin{itemize}

\item[-] By applying $\sigma$ over $a \;\mathbf{at \;least}\; c \;\mathbf{in} \; \{ d_1,...,d_m\}$ we obtain:
\[
    \left(\bigwedge_{i \in C}@_{T_i} a \right)\wedge \left(\bigwedge_{i,j \in C \ \mbox{s.t.} \ i< j}(T_i \ \neq \ T_j)\right) \wedge \left(\bigwedge_{i \in C} \left(\bigvee_{j\in M}(T_i= T-d_j)\right)\right)
\]
where $C = \{1,...,c\}$, $M = \{1,...,m\}$ and $T_1,...,T_n$ are fresh variables.

\item[-] By applying $\sigma$ over $a \;\mathbf{always \; in} \; \{d_1,\dots,d_m\}$ we obtain:
\[
    \left(\bigwedge_{i \in M}@_{T_i} a \right)\wedge \left(\bigwedge_{i,j \in M \ \mbox{s.t.} \ i< j}(T_i \ \neq \ T_j)\right) \wedge \left(\bigwedge_{i \in M} \left(\bigvee_{j\in M}(T_i= T-d_j)\right)\right)
\]
where $M = \{1,...,m\}$ and $T_1,...,T_n$ are fresh variables.


\item[-] By applying $\sigma$ over $a \;\mathbf{count}\; c \;\mathbf{in} \; \{ d_1,\dots,d_m\}$ with $c\in {\cal C}\cap\mathbb{N^+}$ we obtain:
\begin{multline*}
    \left(\bigwedge_{i \in C}@_{T_i} a \right)\wedge \left(\bigwedge_{i,j \in C \ \mbox{s.t.} \ i< j}(T_i \ \neq \ T_j)\right) \wedge \left(\bigwedge_{i \in C} \left(\bigvee_{j\in M}(T_i= T-d_j)\right)\right)\wedge \\
 \neg \left( @_{T_{c+1}}a \ \wedge \ \left(\bigwedge_{i \in C }(T_{c+1} \ \neq \ T_i)\right)\wedge\left(\bigvee_{j\in M}(T_{c+1}= T-d_j)\right)\right)
\end{multline*}
where $C = \{1,...,c\}$, $M = \{1,...,m\}$ and $T_1,...,T_n$ are fresh variables.


\item[-] By applying $\sigma$ over $a \;\mathbf{count}\; v \;\mathbf{in} \; \{ d_1,\dots,d_m\}$ with $v\in {\cal V}$ we obtain the predicate atom $count(b,t_1,\dots,t_n,d_1,\dots,d_m,V)$.

\end{itemize}
Basically, the function $\sigma$ returns a LARS formula that is true when the corresponding $\textit{LDSR}$ streaming atom is true. The formula identifies the time points related to the backward observation of a stream w.r.t the set $\{d_1,...,d_m\}$ for which the condition defined by the streaming atom holds.

\subsection{The function $g$}\label{sec:appendixA} 
We present the function $g$, used by $\rho_4$. In particular, $g$ associates each streaming atom (but those of form $a  \;\mathbf{count}\; v \;\mathbf{in} \; \{ d_1,\dots,d_m\}$) with a $\textit{LARS}_{\textit{D}}$ formula that expresses the condition that must be satisfied in the stream for the streaming atom to be true; it has to consider also the temporary information possibly present in the previous time point when the streaming atom was evaluated. The presence of these information in the stream is simulated with the auxiliary atoms with suffix ``temp''.



\begin{itemize}
    
\item[-] $g(a  \;\mathbf{at \;least}\; 1 \;\mathbf{in} \; \{ d_1,\dots,d_m\})=$ 
\[
atemp \vee  \sigma (a  \;\mathbf{at \;least}\; 1 \;\mathbf{in} \; \{ d_1,\dots,d_m\}\setminus\{0\}).\]

\item[-] Let $c>1$, $g(a  \;\mathbf{at \;least}\; c \;\mathbf{in} \; \{ d_1,\dots,d_m\})=$
\[( atemp\wedge \sigma (a  \;\mathbf{at \;least}\; c-1 \;\mathbf{in} \; \{ d_1,\dots,d_m\}\setminus\{0\}) )\vee \] \[\sigma (a  \;\mathbf{at \;least}\; c \;\mathbf{in} \; \{ d_1,\dots,d_m\}\setminus\{0\})) .\]
\item[-] $g(a  \;\mathbf{always \; in}\;\{ d_1,\dots,d_m\})=$

\[
\begin{cases}
		atemp\wedge \sigma(a  \;\mathbf{always \; in}\; \{ d_1,\dots,d_m\}\setminus\{0\} & \text{if $0 \in \{d_1,\dots,d_m\}$}\\
        \sigma(a \;\mathbf{always \; in}\; \{ d_1,\dots,d_m\}) & \text{otherwise}
		 \end{cases}\]

\item [-]$g(a  \;\mathbf{count}\; 1 \;\mathbf{in} \; \{ d_1,\dots,d_m\})=$
\[atemp \vee  ( \neg atemp \wedge \sigma (a  \;\mathbf{count}\; 1 \;\mathbf{in}  \{ d_1,\dots,d_m\}\setminus\{0\})).\]

\item [-]Let $c>1$, $g(a  \;\mathbf{count}\; c \;\mathbf{in} \; \{ d_1,\dots,d_m\})=$
\begin{multline*}
    (atemp \wedge \sigma(a  \;\mathbf{count}\; c-1 \;\mathbf{in} \; \{ d_1,\dots,d_m\}\setminus\{0\})) \vee \\ \sigma(a  \;\mathbf{count}\; c \;\mathbf{in} \; \{ d_1,\dots,d_m\}\setminus\{0\}).
\end{multline*}

\end{itemize}
Moreover, if $\alpha$ is a streaming atom of the form $a \;\mathbf{count}\; v \;\mathbf{in} \; \{ d_1,\dots,d_m\}$ or an aggregate atom, $g$ associates it with a formula containing auxiliary atoms defined via an additional set of rules $C_{\alpha}$ that are needed to simulate its semantics.
Let $C\in {\cal V}$, $g(a  \;\mathbf{count}\; C \;\mathbf{in} \; \{ d_1,\dots,d_m\})=$
\[( atemp\wedge \sigma(a  \;\mathbf{count}\; C_1 \;\mathbf{in} \; \{ d_1,\dots,d_m\}\setminus\{0\})\wedge (C_1+1=C) \\ \vee \] \[(\neg atemp\wedge \sigma(a  \;\mathbf{count}\; C \;\mathbf{in} \; \{ d_1,\dots,d_m\}\setminus\{0\}))  \]
The set of rules $C_{\alpha}$ defining $\alpha=b(t_1,\dots,t_n)  \;\mathbf{count}\; v \;\mathbf{in} \; \{ d_1,\dots,d_m\}$ where $v\in V$ are the following $m+1$ rules: 
\begin{itemize}
\item $present (b,t_1,\dots,t_n,d_1,\dots,d_m,c)\leftarrow \sigma(b(t_1,\dots,t_n)  \;\mathbf{at \;least}\; c \;\mathbf{in} \; \{ d_1,\dots,d_m\})$   
for $c\in{1,...,m}$

\item $count (p,t_1,\dots,t_n,d_1,\dots,d_m,C)\leftarrow present (b,t_1,\dots,t_n,d_1,\dots,d_m,C), \\ \neg present (b,t_1,\dots,t_n,d_1,\dots,d_m,C+1)$.
\end{itemize}
First, $m$ rules are needed, where each one verifies if atoms of type $b(t_1,\dots,t_n)$ occur $1,2,...,$ or $m$ times. Eventually, a rule is needed to verify the exact number of occurrences.
Similar approaches are used for the translation of the aggregate atoms. \\
Eventually, $g(\mathtt{not}\; \alpha)=\neg   g( \alpha)$ for each positive literal $\alpha$.

\subsection{The function $g'$}\label{sec:appendixB}
We present the function $g'$, used by $\rho_5$. In particular, $g'$ associates each streaming atom (but those of form $a  \;\mathbf{count}\; v \;\mathbf{in} \; \{ d_1,\dots,d_m\}$) with a $\textit{LARS}_{\textit{D}}$ formula that expresses the condition that must be satisfied in the stream for the streaming atom to be true.
\begin{itemize}
\item[-]$g'(a \;\mathbf{at \;least}\; c \;\mathbf{in} \; \{ d_1,\dots,d_m\})= \sigma(a  \;\mathbf{at \;least}\; c \;\mathbf{in} \; \{ d_1,\dots,d_m\})$.

\item[-]$g'(a \;\mathbf{always \; in} \; \{d_1,\dots,d_m\})=\sigma(a  \;\mathbf{always \; in} \; \{d_1,\dots,d_m\})$.

\item[-]$g'(a  \;\mathbf{count}\; k \;\mathbf{in} \; \{ d_1,\dots,d_m\})= \sigma (a  \;\mathbf{count}\; k \;\mathbf{in} \; \{ d_1,\dots,d_m\})$ if $k\in \mathbb{N^+}$.

\end{itemize}
Moreover, if $\alpha$ is a streaming atom of the form $a  \;\mathbf{count}\; v \;\mathbf{in} \; \{ d_1,\dots,d_m\}$ or an aggregate atom, $g'$ associates it with a formula containing auxiliary atoms defined via an additional set of rules $C_{\alpha}$ that are needed to simulate its semantics. 
Let $v\in V$, $g'( b(t_1,\dots,t_n) \;\mathbf{count}\; v \;\mathbf{in} \; \{ d_1,\dots,d_m\}) = count(b,t_1,\dots,t_n,d_1,\dots,d_m,V)$. 
The set of rules $C_{\alpha}$ defining $\sigma(\alpha)$ for $\alpha=b(t_1,\dots,t_n)  \;\mathbf{count}\; v \;\mathbf{in} \; \{ d_1,\dots,d_m\}$ where $v\in V$ are the following $m+1$ rules: 
\begin{itemize}
\item $present (b,t_1,\dots,t_n,d_1,\dots,d_m,c)\leftarrow \sigma(b(t_1,\dots,t_n)  \;\mathbf{at \;least}\; c \;\mathbf{in} \; \{ d_1,\dots,d_m\})$   
for $c\in{1,...,m}$

\item $count (b,t_1,\dots,t_n,d_1,\dots,d_m,C)\leftarrow present (b,t_1,\dots,t_n,d_1,\dots,d_m,C), \\ \neg present (b,t_1,\dots,t_n,d_1,\dots,d_m,C+1)$.
\end{itemize}
First, $m$ rules are needed, where each one verifies if atoms of type $b(t_1,\dots,t_n)$ occur $1,2,...,$ or $m$ times. Eventually, a rule is needed to verify the exact number of occurrences. Similar approaches are used for the translation of the aggregate atoms.

Eventually, $g'(\mathtt{not}\; \alpha)=\neg   g'( \alpha)$ for each positive literal $\alpha$.

\subsection{The function $g''$}\label{sec:appendixC}
We present the function $g''$, used by $\rho_7$. In particular, $g''$  associates each streaming atom (but those of form $a \;\mathbf{count}\; v \;\mathbf{in} \; \{ d_1,\dots,d_m\}$) with a $\textit{LARS}_{\textit{D}}$ formula that expresses the condition that must be satisfied in the stream for the streaming atom to be true; it has to consider the temporary information possibly present in a time point $t$ when $t$ was the evaluation time point and thus the streaming atom was evaluated. It is possible to verify the presence of an atom $a$ at the time of a previous evaluation also if was derived by a rule of form (2): since in the body of any rule of form (2) are only present atoms deriving from the input stream o from rules of form (1), we can know if an atom was derived considering its derivation $d_P(a)$.
We assume that the program rules do not share variables. 
\begin{itemize}

\item[-] $\sigma(a \vee b)= \sigma(a) \vee \sigma(b) $. 
\item[-] $\sigma(a\wedge b)= \sigma(a) \wedge \sigma(b)$.
\item [-]$\sigma(\mathtt{not}\; \alpha)=\neg   \sigma( \alpha)$

\item[-] $g^{''}(a  \;\mathbf{at \;least}\; 1 \;\mathbf{in} \; \{ d_1,\dots,d_m\})=$ 
\[\sigma( d_P(a))  \vee \sigma (a  \;\mathbf{at \;least}\; 1 \;\mathbf{in} \; \{ d_1,\dots,d_m\}\setminus\{0\})\]

\item[-] Let $c>1$, $g{''}(a \;\mathbf{at \;least}\; c \;\mathbf{in} \; \{ d_1,\dots,d_m\})=$ \[ (g( d_P(a)\wedge g (a  \;\mathbf{at \;least}\; c-1 \;\mathbf{in} \; \{ d_1,\dots,d_m\}\setminus\{0\}))  \vee \] \[ g (a  \;\mathbf{at \;least}\; c \;\mathbf{in} \; \{ d_1,\dots,d_m\}\setminus\{0\})) \]

\item[-] $g''(a  \;\mathbf{always \; in}\;\{ d_1,\dots,d_m\})=$

\[
\begin{cases}
	\sigma(d_P(a))\wedge \sigma(a  \;\mathbf{always \; in}\; \{ d_1,\dots,d_m\}\setminus\{0\} & \text{if $0 \in \{d_1,\dots,d_m\}$}\\
        \sigma(a \;\mathbf{always \; in}\; \{ d_1,\dots,d_m\}) & \text{otherwise}
		 \end{cases}\]

\item [-]$g''(a  \;\mathbf{count}\; 1 \;\mathbf{in} \; \{ d_1,\dots,d_m\})=$
\[	\sigma(d_P(a)) \vee  ( \neg 	\sigma(d_P(a)) \wedge \sigma (a  \;\mathbf{count}\; 1 \;\mathbf{in}  \{ d_1,\dots,d_m\}\setminus\{0\})).\]

\item [-]Let $c>1$, $g''(a  \;\mathbf{count}\; c \;\mathbf{in} \; \{ d_1,\dots,d_m\})=$
\begin{multline*}
    	\sigma(d_P(a)) \wedge \sigma(a  \;\mathbf{count}\; c-1 \;\mathbf{in} \; \{ d_1,\dots,d_m\}\setminus\{0\})) \vee \\ \sigma(a  \;\mathbf{count}\; c \;\mathbf{in} \; \{ d_1,\dots,d_m\}\setminus\{0\})).\
\end{multline*}

\end{itemize}

Eventually, $g''(\mathtt{not}\; \alpha)=\neg   g''( \alpha)$ for each positive literal $\alpha$. 
\section{Proofs}\label{sec:appendixP} 

\subsection{Statement and Proof of Lemma 1}\label{sec:appendixL}
\begin{lemma}\label{until}
Let $P$ be a \textit{LDSR} program, $I=\langle I_0, \dots, I_n \rangle$ an input stream and $t\in \{1,\dots,n\}$, for all set of ground predicate atoms $B\subset G$ and for all $I'=\langle I'_0, \dots, I'_n \rangle$ with $I_i=I'_i$ for $i=0,\dots,t$, we have that $t-bound(I,B,P)=t-bound(I',B,P)$.
\end{lemma}

Let  $t-bound(I,B,P)=O=\langle O_{0},\dots,O_{n}\rangle$ and  $t-bound(I',B,P)=O'=\langle O'_{0},\dots,O'_{n} \rangle$, for definition of answer stream:
\begin{itemize}
\item [-] $\langle O_0\rangle$ is the permanent part of a minimal model $M$ of the reduct ${P^{+}}^{M}$ for the stream $\langle I_0 \rangle$ and $\langle O'_0\rangle$ is the permanent part of a minimal model $M$ of the reduct ${P^{+}}^{M}$ for the stream $\langle I'_0 \rangle$. Since $I_0=I'_0$ we have that $O_0=O'_0$;
\item [-]  for all $i \in 1,\dots,t-1$, $O_{|i}$ is the permanent part of a minimal model $M$ of the reduct ${P^{+}}^{M}$ for the stream $\langle O_{0},\dots,O_{i-1}, I_{i} \rangle$. Since $I_i=I'_i$, we have that $O_{|{i}}=O'_{|{i}}$.
\item [-] $O_{|t}$ is a minimal model of the reduct ${P^{+}}^{O_{|t}}$ for the stream $\langle O_0,\dots,O_{t-1}, I_t \rangle$. Since $I_t=I'_t$, we have that $O_{|{t}}=O'_{|{t}}$.
\end{itemize}
Finally, for definition $O_i=O'_i=\emptyset$ for $t < i \leq n$.

\subsection{Proof of Proposition 1}\label{sec:appendixP1}

Let $P$ be the LARS program $\{ @_{T-1}\; a \leftarrow \; @_T c\}$.

Let $T = \{0,\dots,n\}$ and $0<\tau<n$.
Let $C \subseteq T \setminus \{0\}$ any subset of T such that $\tau+1 \in C$ and $i \in C \rightarrow i-1 \not\in C$. Let $A = \{i-1 | i \in C\}$. Let $I = \langle I_0, I_1, \dots,I_n\rangle$ a stream such that:
\\$I_i= 
\begin{cases} 
\emptyset & \mbox{if } i \not\in C\\
\{c\} & \mbox{if } i \in C\\
\end{cases}$ 

Let $O=\langle O_0, O_1, \dots,O_n\rangle \in AS(P,I,\tau)$. Observe that:
\begin{itemize}
\item [-]  $O_i = \{a\}$ if $i \in A$
\item [-]  $O_i = \{c\}$ if $i \in C$
\item [-]  $O_i = \emptyset$ if $i \not\in (A \cup C)$
\end{itemize}
 Assume, by contradiction, that there exists a mapping $\rho: LARS \rightarrow \textit{LDSR}$ such that, for each $I$ and $B$ and for each time point of evaluation $t \in \{0,\dots,n\}$, it holds that 
$t-atomic(I,B,P)=t-atomic(I,B,\rho(P))_{|_{pred(P\cup I\cup B)}}$. 
Let $B=\emptyset$ and $I' = \langle I_0', I_1', \dots,I_n'\rangle$ a stream such that:
\\$I_i'= 
\begin{cases} 
I_i & \mbox{if } i \neq \tau+1 \\
\emptyset & \mbox{if } i = \tau+1\\
\end{cases}$ 

Let $\rho(P)=\bar{P}$ and $O'=\langle O'_0, O'_1, \dots,O'_n\rangle \in AS(P,I',\tau)$, we have that $O_{\tau} = a$ and $O'_{\tau} = \emptyset$.\\
 Since for the Lemma~\ref{until} $\tau-bound(I,B,\bar{P})_{|_\tau} = \tau-bound(I',B,\bar{P})_{|_\tau}$, we have that \\ $\tau-atomic(I',B,P) \neq \tau-atomic(I',B,\bar{P})$.

\subsection{Proof of Proposition 2}\label{sec:appendixP2}

Let $P$ be the \textit{LDSR} program $\{a(Y) \derives a(X), b(X,Y).\}$
Let $T = \{0,\dots,n\}$ and $\tau>0\in T$. Let $I = \langle I_0, I_1, \dots,I_n\rangle$ a stream with $I_i=\{a(1),b(1,2)\}$ for $i\in \{0,\dots,n\}$.
Assume, by contradiction, that there exists a mapping $\rho: \textit{LDSR} \rightarrow LARS$ such that, for each $I$ and $B$ and for each time point of evaluation $t \in \{0,\dots,n\}$, it holds that 
$t-atomic(I,B,P)=t-atomic(I,B,\rho(P))_{|_{pred(P\cup I\cup B)}}$.

Let $B=\emptyset$, $\rho(P)=\bar{P}$ and $\tau-atomic(I,B,P)=\tau-atomic(I,B,\bar{P})=\langle O_0, O_1, \dots,O_n\rangle$, we have that $O_{\tau} = \{a(1),a(2),b(1,2)\}$.\\
For definition, let $A=\langle A_0,\dots, A_n\rangle\in AS(P,I,t)$, we have that $O_\tau=A_\tau\cup B$, but $A$ is an interpretation stream for $I$, so all atoms that occur in $A$ but not in $I$ must have intensional predicates but the atom $a(2)$ has predicate $a$ with $a\in P_{S}^{\varepsilon}$.

\end{document}